# DUAL-NEIGHBORHOOD DEEP FUSION NETWORK FOR POINT CLOUD ANALYSIS

*Guoquan Xu*[*], *Hezhi Cao*[†], *Yifan Zhang*[*], *Jianwei Wan*[*], *Ke Xu*[*] and *Yanxin Ma*[*]

[*]National University of Defense Technology, Changsha, CHINA
[†] University of Science and Technology of China, Hefei, CHINA
{xuguoquan19, zhangyifan16c, xuke, mayanxin}@nudt.edu.cn,
caohezhi21@mail.ustc.edu.cn, kermitwjw@139.com

## ABSTRACT

Recently, deep neural networks have made remarkable achievements in 3D point cloud analysis. However, the current shape descriptors are inadequate for capturing the information thoroughly. To handle this problem, a feature representation learning method, named Dual-Neighborhood Deep Fusion Network (DNDFN), is proposed to serve as an improved point cloud encoder for the task of point cloud analysis. Specifically, the traditional local neighborhood ignores the long-distance dependency and DNDFN utilizes an adaptive key neighborhood replenishment mechanism to overcome the limitation. Furthermore, the transmission of information between points depends on the unique potential relationship between them, so a convolution for capturing the relationship is proposed. Extensive experiments on existing benchmarks especially non-idealized datasets verify the effectiveness of DNDFN and DNDFN achieves the state of the arts.

## 1. INTRODUCTION

In recent years, 3D point cloud analysis has received quantities of attention since the disclosure of many datasets. 3D point cloud classification is one of the hottest research directions, for its important application value in many fields [1]. Thus, the approaches to effectively and efficiently processing 3D point clouds are critically needed and researchers have proposed a series of innovative approaches and achieved satisfied results.

Previous studies can be roughly divided into three categories: projection, voxelization, and point-based methods. The projection [2, 3] usually causes information loss. Even worse, this method often requires a large number of views to obtain better performance, which is almost impossible in real scenarios (only one view can be obtained in practical). Similarly, the voxelization [4, 5] induces a loss of information, excessive consumption of memory, and high computation cost.

In contrast to the above two categories of methods, the current mainstream methods aim to deal with point clouds directly. As a pioneering work, PointNet [6] proposes to learn the spatial encodings of points by combing Multi-Layer Perceptron (MLP) and global aggregation. Then, the follow-up work, PointNet++ [7] explores local structure information aggregation to improve the network. However, all points are processed with the same MLP, which still limits the feature representation ability of the network. Therefore, most recent works [8-11] have attempted to design operations similar to convolution to capture spatial correlation.

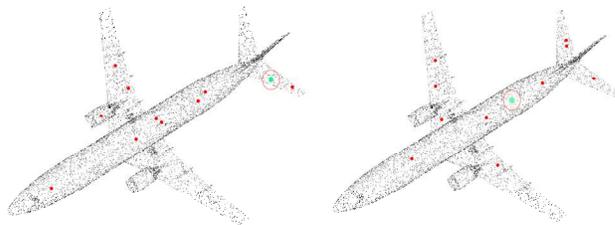

**Fig. 1.** Adaptive key neighborhood search. The green point is the central point. It has a local neighborhood (red circle part) and the corresponding key points (red points) can be adaptively searched through our method. Each local information has its unique key points.

These methods adopt the "local-to-global" mode which extracts the local structure information, and obtain the global information through global pooling. However, too much focus on local information will confines the receptive field. Injecting global key information into local information will enable the network to broaden the eyesight for each local point patch. Hence, a method called Trainable-Neighborhood Learning (TN-Learning) is proposed in this paper to search the global key information. As shown in Fig. 1, the key information corresponding to each local information is different and TN-Learning can adaptively search the matching information. At the same time, a deep fusion network is designed to inject key information.

Information aggregation is the key part of feature extraction. Each pair of points has a unique potential relationship, which is ignored by many methods. Therefore, Information Transfer Convolution (IT-Conv) is proposed to mine this potential relationship as a criterion for information aggregation. The main contributions of this paper are summarized as follows:

- A global key neighborhood learning method is illustrated as TN-Learning, which adaptively searches the key neighborhood as a supplement to kNN or ball query to improve the performance of the network.

- A dual neighborhood feature extraction network structure is designed to extract features from the key neighborhood and local neighborhood respectively, and the features are combined to improve the representation ability of the network.
- An efficient convolution operation named IT-Conv is designed. It determines aggregation rules by learning the unique potential relationship between each pair of points.
- Extensive experiments are exhibited on challenging benchmarks across three tasks including real-world benchmarks, demonstrating DNDFN achieves the state of the arts.

## 2. RELATED WORK

### 2.1. Point-based Deep Learning

PointNet [6] pioneers point cloud learning by using MLP, max-pooling, and rigid transformations to ensure invariance under permutations and rotation. Subsequently, PointNet++ [7] uses ball query grouping and hierarchical PointNet to capture local structures. Inspired by them, further works aim to design convolution operations on point clouds. DGCNN [8] designs EdgeConv which captures similar local shapes by learning point relation in a high-dimensional feature space. However, this relation is unreliable in some cases. Moreover, DGCNN recalculates the nearest point through kNN at each layer to obtain the dynamic graph. RS-Conv [9] is proposed for RS-CNN as a learn-from-relation convolution operator. RS-Conv learns a high-level relation expression from geometric priors in 3D space. It can explicitly encode geometric relation of points, thus resulting in much shape awareness and robustness.

### 2.2. Neighborhood Selection

The most commonly used neighborhood selection methods are kNN and ball query. The former takes the $k$ points closest to the central point as the neighborhood and the latter finds all points that are within a radius to the query point. A multi-directional search method has been proposed in Point Attention Network [10] which divides the ball into $k$ bins along the azimuth. In each bin, $m$ points closest to the central point have been taken as the neighborhood. This method ensures that the neighborhood points come from different directions, so the local shape can be better represented. In addition, there are many different variants. However, these methods are essentially based on distance relationships and focus on the local structure information. Thus TN-Learning is proposed to supplement the global key information in this paper. It is quite effective to use global key information to assist local information in reasoning, and is also in line with the way of thinking of the human brain.

### 2.3. Fusion Network

Fusion Network is an efficient network type and can be divided into two types: feature fusion and decision fusion. DPDFN [12] learns the key and local features of the human face through GMN and LRN sub-networks respectively and it is a decision fusion network. Zhang et al. [13] proposes to aggregate point features and voxel features respectively through two branches. PointFusion [14] extracts and fuses the 3D point cloud features and RGB features respectively through PointNet [6] and ResNet. Based on these works, a feature fusion network that can extract local structure information and corresponding key information by two branches is designed in this paper.

## 3. METHOD

The cores of DNDFN are TN-Learning, IT-Conv, and fusion network. TN-Learning adaptively searches the corresponding key neighborhood and then IT-Conv is responsible for transmitting information between the central point and its neighbors. Finally, local features and corresponding key features are fused by the fusion network. This section orderly introduces the design methods of these three points.

### 3.1. TN-Learning

Adding key information to kNN or ball query can make the features extracted by convolution stronger. The key information corresponding to each local information is different, so an adaptive search method TN-Learning is proposed.

Given a point cloud $X = \{x_i | i=1,2,…,N\} \in \mathbb{R}^{N \times 3}$, and it has features $F = \{f_i | i=1,2,…N\} \in \mathbb{R}^{N \times C}$. $N$ is the number of points and $C$ is the dimension of features. Sample the point $x_i$ as the central point and it has $2k$ neighbors $\mathcal{N}(x_i)$. These neighbors are divided into local neighbors $\mathcal{N}_{loc}(x_i)$ and key neighbors $\mathcal{N}_{key}(x_i)$. $\mathcal{N}_{loc}(x_i)$ can be obtained by kNN or ball query. Ball query is chosen in this paper. $\mathcal{N}_{key}(x_i)$ is generated by TN-Learning. TN-Learning learns the relationship coefficients between points and this process can be formulated as:

$$\xi_{ij} = \langle \phi(f_i), \phi(f_j) \rangle, \forall x_j \in X \text{ and } x_j \neq x_i, \quad (1)$$

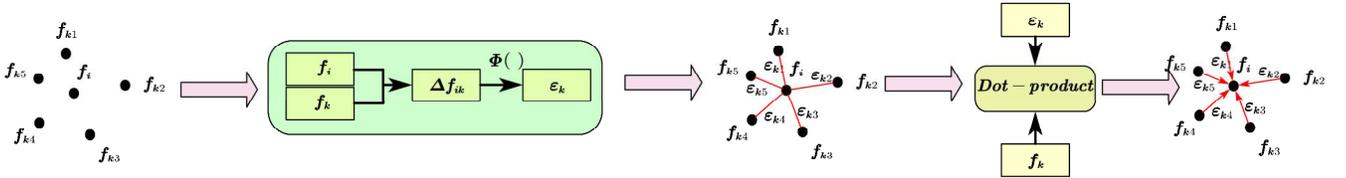

**Fig. 2.** Neighborhood information is transmitted to the central point by unique transfer rules. $\Delta f_{ik}$ is used to learning the information of relationship $\varepsilon_k$. That is, each pair of points is defined with unique information transmission rules. And then adjacent vertex feature $f_k$ transmitted to the central point based on the rule $\varepsilon_k$.

where $\xi_{ij}$ is the relationship coefficient between $x_i$ and $x_j$. $\langle , \rangle$ represents dot product. Operation $\phi$ attempts to learn the descriptors of features and can be described as follows:

$$\phi(f_i) = \Pi\left(\sigma(\Theta(f_i))\right), \quad (2)$$

where both $\Pi$ and $\Theta$ are mapping functions and implemented with MLP. $\sigma$ is the activation function LeakyReLU. The relationship coefficients between the central point and other points can be obtained by Eq. (1). The coefficient value represents the strength of the relationships, and $k$ points with the strongest relationships can be selected as neighbors.

The local neighborhood is constant, and the key neighborhood can be adjusted according to the loss of each training. Each local neighborhood has corresponding and unique key neighborhood.

### 3.2. IT-Conv

Ball query obtains local information, while TN-Learning searches for key information from the global, and IT-Conv further learns the potential relationship between neighborhood and the central points. This relationship determines the way information is transmitted. Thus IT-Conv can be described as follows:

$$f_i' = \alpha\left(\mathcal{A}(\{\Phi(f_k - f_i) \cdot f_k, \forall x_k \in \mathcal{N}(x_i)\})\right), \quad (3)$$

where $\alpha$ is a nonlinear activator and $\mathcal{A}$ is an aggregate function. The mapping function $\Phi$ is used to learn the relationship between the central point and its neighbors. Eq. (3) indicates that the information of neighbors is transmitted to the central point through learned relationships. Compared with the simple aggregation of information, this kind of information aggregation based on the learned rules is obviously more effective. This process can be represented by Fig. 2. The features of two points are used to learn their relationship $\varepsilon_k$, which then determines the rules of information transmission.

TN-Learning evaluates the importance of each key neighbor. Further, the evaluation result, *i.e.*, relationship coefficient $\xi$, can be used to weight in the process of information aggregation. Thus, a coefficient is designed for information transmission in the key neighborhood as:

$$\omega_k = \frac{\exp(\xi_{ik})}{\sum_{x_l \in \mathcal{N}_{key}(x_i)} \exp(\xi_{il})}. \quad (4)$$

As a result, IT-Conv can be improved to weighted IT-Conv:

$$f_i'' = \alpha\left(\mathcal{A}(\{\omega_k \cdot \Phi(f_k - f_i) \cdot f_k, \forall x_k \in \mathcal{N}(x_i)\})\right) \quad (5)$$

Consequently, the local neighborhood information is transmitted using IT-Conv, while the key neighborhood information is transmitted according to weighted IT-Conv. This can fully mine the potential information between each pair of points, so it can better aggregate information.

### 3.3. Dual-Neighborhood Fusion Network

Key information is the supplement of local information, that is, they need information fusion to get strong information. Therefore, a feature fusion network is designed to fuse local neighborhood and key neighborhood. The classification network of DNDFN is designed as shown in Fig. 3. It consists of four convolution layers and two fully connected (FC) layers. The core part of this network is the Dual-Neighborhood Fusion Encoder (DNFE). The input is sent into two branches to extract local features and key features respectively, and then the features are concatenated directly together. The output of this layer is finally obtained through an MLP. Conv1 corresponds to IT-Conv, while Conv2 corresponds to weighted IT-Conv. The parameter settings of the two branches are consistent. In order to make the network stronger, RS-Conv [9] based on ball query is employed to extract all features point by point in the first layer. Subsequently, DNFE is implemented on the next three layers.

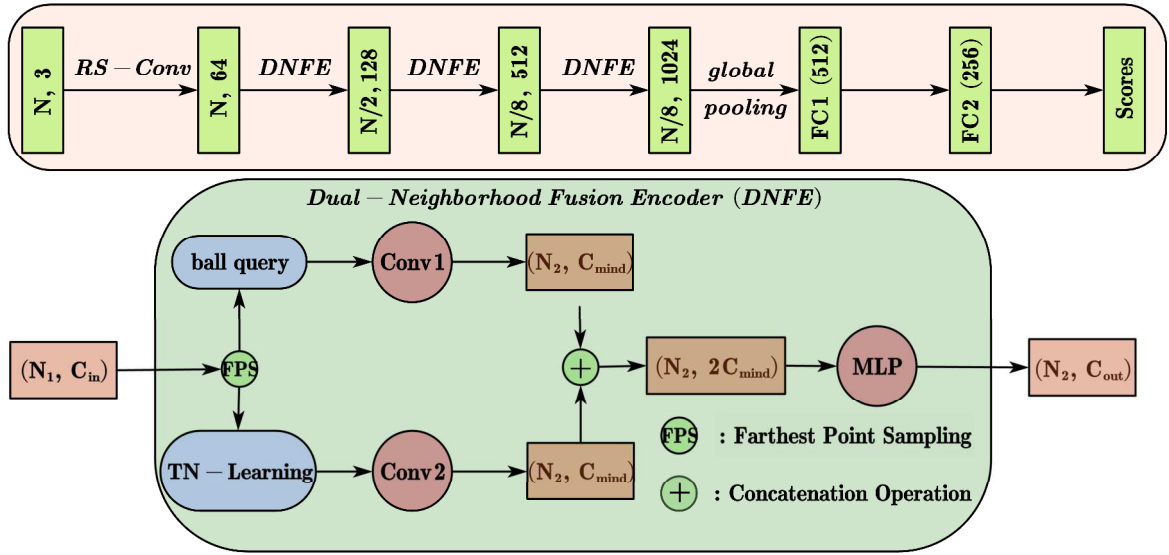

**Fig. 3.** The classification network of DNDFN. DNDFN has four convolution layers and two full connected layers. RS-Conv with ball query is employed in the first layer. Then, DNFE is applied to the next three layers. Conv1 corresponds to the convolution operation of IT-Conv, while Conv2 corresponds to weighted IT-Conv. The upper branch is responsible for capturing local information, while the lower branch searches for corresponding key information. The features of the two branches are directly merged together.

## 4. EXPERIMENTS

In this section, comprehensive experiments are arranged to validate the proposed DNDFN. First of all, DNDFN is evaluated on idealized datasets ModelNet40 [15] and ShapeNet [16]. Then the non-idealized dataset ScanObjectNN [17] is employed to the experiments. Finally, ablation studies and robustness experiments are implemented to verify the effectiveness of the modules and the performance of the network. It is worth mentioning that the network structure and parameters are consistent for all classification tasks except for necessary changes.

### 4.1. Shape Classification on ModelNet40

ModelNet40 [15] classification benchmark contains 9,843 train models and 2,468 test models in 40 classes. PointNet [6] samples point cloud data from them and then 1024 points are sampled uniformly in this paper. Voting tests with random scaling and average the predictions are performed during testing. The main parameter settings are as follows: the rate of dropout is set to 50% in the last two fully-connected (FC) layers; batch normalization and LeakyReLU are applied on all layers; the SGD optimizer with momentum set to 0.9 is adopted; The initial learning rate is donated to 0.1 and is dropped to 0.001 by cosine annealing. If there are no special instructions, these parameter settings are also adopted for other subsequent experiments.

The results of DNDFN and the state-of-the-art methods with only the original coordinates used as input are shown in Table 1. DNDFN ranks second, which is a very competitive result. Even without voting, DNDFN still achieves 92.9% ("no vote" indicates without voting tests).

To further illustrate the compensation mechanism of TN-Learning, the neighbors searched in the second and third layers of the network are visualized in Fig. 4. Donating pink to indicate the central point. The neighbors searched by ball query and our method are represented by green and blue respectively. Red indicates the neighbors repeatedly searched by the two methods. As shown in the figure, TN-Learning captures the long-distance neighborhood that ball query cannot capture. This process is fully adaptive. From the second layer to the third layer, the blue points gradually fill the whole object (except the part searched by ball), and green, blue and red points gradually depict the main body of the object. Moreover, the most useful part for classification is the edge for flat objects, such as desktop. The rightmost two pictures show that blue points are distributed on the edge of the desktop. This means that our method can not only adaptively learn the compensation neighborhood, but also learn the key parts.

**Table 1.** Classification accuracy (%) on ModelNet40.

| Method | input | Accuracy |
| --- | --- | --- |
| PCNN [10] | 1K points | 92.3 |
| FPConv [18] | 1K points | 92.5 |
| PointCNN [11] | 1K points | 92.5 |
| PointASNL [19] | 1K points | 92.9 |
| KPConv [20] | 1K points | 92.9 |
| DGCNN [8] | 1K points | 92.9 |
| DensePoint [21] | 1K points | 93.2 |
| PosPool [22] | 5K points | 93.2 |
| PCT [23] | 1K points | 93.2 |
| RS-CNN [9] | 1K points | **93.6** |
| DNDFN (no vote) | 1K points | 92.9 |
| DNDFN | 1K points | 93.2 |

### 4.2. Shape Part Segmentation on ShapeNet

ShapeNet [16] is composed of 16,881 3D models. It has 16 categories and is labeled in 50 parts in total. As in PointNet [6], each model is downsampled to 2048 points as input.

**Table 2.** Shape part segmentation results (%) on ShapeNet.

| Method | Class mIoU | Instance mIoU |
| --- | --- | --- |
| PCNN [10] | 81.8 | 85.1 |
| PointNet++ [7] | 81.9 | 85.1 |
| DGCNN [8] | 82.3 | 85.1 |
| SpiderCNN [24] | 81.7 | 85.3 |
| SPLATNet [25] | 83.7 | 85.4 |
| PointConv [26] | 82.8 | 85.7 |
| PointCNN [11] | 84.6 | 86.1 |
| RS-CNN [9] | 84.0 | **86.2** |
| DNDFN | 83.1 | 86.0 |

Table 2 summarizes the quantitative comparisons with the state-of-the-art methods. Two types of mean Inter-over-Union (mIoU), class mIoU and instance mIoU, are reported to evaluate the performance of segmentation. The results show that both class mIoU and instance mIoU of DNDFN are very satisfied. Visual segmentation results are shown in Fig. 5.

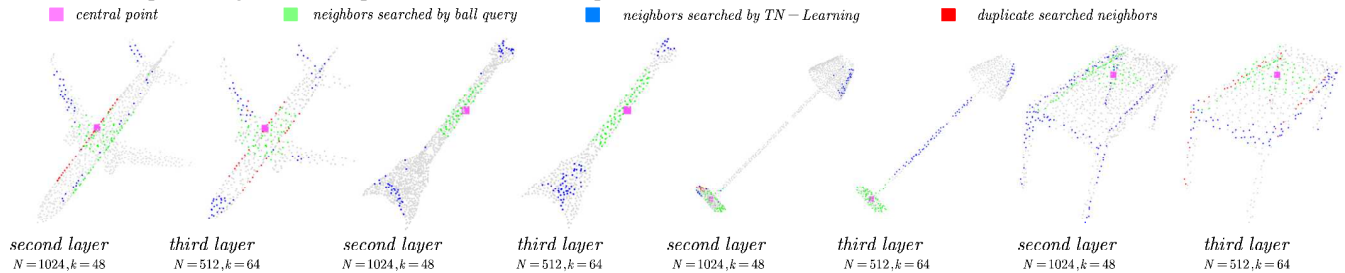

**Fig. 4.** Visualization of searched neighbors. The pink one indicates the central point. The green, blue, and red points represent the neighbors searched by ball query, TN-Learning, and overlapped by ball query and TN-Learning respectively. It can be seen that TN-Learning adaptively searches the compensation neighborhood for ball query and attempts to form the main structure of the object together with ball query. In addition, the blue points are mainly concentrated on the edge of the object, which is the part with the highest degree of object recognition. This shows that our method pays more attention to the key parts when searching adaptive compensation neighbors.

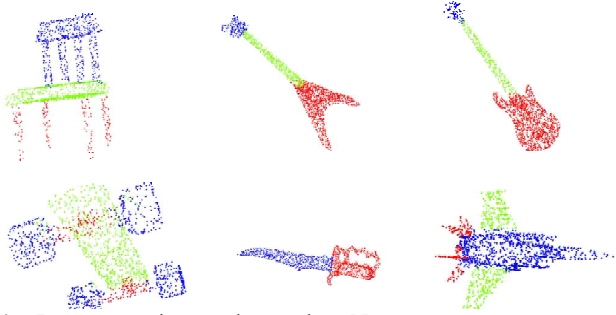

**Fig. 5.** Segmentation results on ShapeNet.

### 4.3. Shape Classification on ScanObjectNN

DNDFN is further applied on ScanObjectNN [17] to evaluate the performance in this subsection. ScanObjectNN is a recent real-world point cloud object dataset based on scanned indoor scene data. The dataset includes three subsets: OBJ_ONLY, OBJ_BG, and HARDEST. OBJ_ONLY includes only ground truth and OBJ_BG adds additional background data. Then, OBJ_BG is extended by translating, rotating (around the gravity axis), and scaling the ground truth bounding box to obtain the HARDEST.

DNDFN is implemented on three subsets respectively, and the results are shown in Table 3. It can be seen that DNDFN has achieved the state of the arts on all three subsets. The same results have been obtained on OBJ_ONLY and OBJ_BG, which shows that DNDFN can be unaffected by the background.

**Table 3.** Classification accuracy (%) on ScanObjectNN.

| Method | HARDEST | OBJ_BG | OBJ_ONLY |
| --- | --- | --- | --- |
| PointNet [6] | 68.2 | 73.3 | 79.2 |
| SpiderCNN [24] | 73.7 | 77.1 | 79.5 |
| PointNet++ [7] | 77.9 | 82.3 | 84.3 |
| RS-CNN [9] | 78.0 | 85.7 | 85.5 |
| DGCNN [8] | 78.1 | 82.8 | 86.2 |
| PointCNN [11] | 78.5 | 86.1 | 85.5 |
| AdaptConv [27] | 78.9 | 84.9 | 84.3 |
| BGA-DGCNN [17] | 79.7 | - | - |
| BGA-PN++ [17] | 80.2 | - | - |
| DNDFN | **81.4** | **87.8** | **87.8** |

### 4.4. Ablation Studies and Robustness Experiments

Detailed ablation studies and robustness experiments on DNDFN are performed to evaluate the performance of our designs.

First of all, the performance of TN-Learning and IT-Conv on the HARDEST dataset is evaluated in Table 4. IT-Conv only based on TN-Learning, ball query and kNN are competitive with accuracies of 79.4%, 80.3% and 80.4%, respectively. In addition, the results of their pairwise combination are also reported. The results always improve when there is TN-Learning, while the fusion result of ball query and kNN decrease slightly. This is because TN-Learning can compensate adaptively, while the combination of ball query and KNN produces information redundancy.

**Table 4.** The performance of different modules of DNDFN.

| TN-Learning | Ball query | kNN | Accuracy (%) |
| --- | --- | --- | --- |
| √ | | | 79.4 |
| | √ | | 80.3 |
| | | √ | 80.4 |
| | √ | √ | 80.2 |
| √ | | √ | 81.2 |
| √ | √ | | 81.4 |

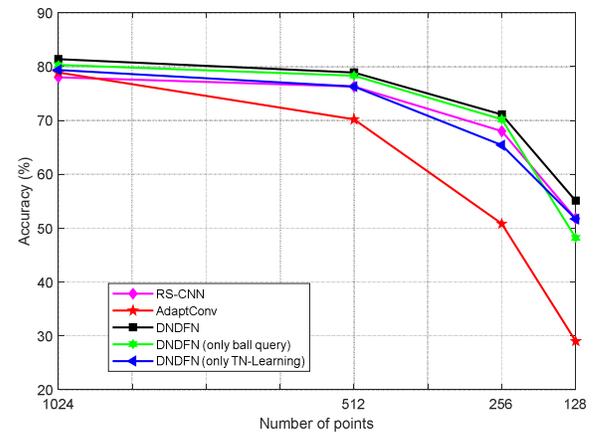

**Fig. 6.** The results of robustness experiments.

In the robustness experiments, DNDFN trained with 1024 points is respectively tested with sparser points of number 128, 256, 512, and 1024 as the input. In order to be more convincing, the experiments are operated on the HARDEST subset. As can be seen in Fig. 6, compared with 1024 points as input, the result decreased only 3.1% and 12.7% on the input of 512 and 256 points, which illustrates the robustness of DNDFN.

**Table 5.** Classification accuracy (%) on rotated ModelNet40

| Method | original | arbitrary rotation |
| --- | --- | --- |
| PointNet [6] | 89.2 | 75.5 (13.7 ↓) |
| PointNet++ [7] | 90.7 | 77.4 (13.3 ↓) |
| DGCNN [8] | 92.9 | 81.1 (11.8 ↓) |
| RS-CNN [9] | **93.6** | 84.6 (9 ↓) |
| DNDFN | 93.2 | **87.6 (5.6 ↓)** |

Rotation is one of the most common problems in realistic. The performance of existing works is greatly reduced under the influence of rotation. As discussed before, DNDFN performs well on the HARDEST which contains the rotated data. In order to further verify its performance against the rotation problem, DNDFN is trained and tested on arbitrarily rotated ModelNet40. The parameters and network structure of DNDFN remain unchanged. As shown in Table 5, DNDFN performs significantly better.

In order to compare the complexity of our method with the previous method, the relevant results in Table 6. From the table, it can be seen that our model achieves the best performance of 93.2 % overall accuracy and the model size is relatively small.

**Table 6.** The number of parameters and overall accuracy of different methods

| Method | #parameters | Accuracy (%) |
| --- | --- | --- |
| PointNet [6] | 3.5M | 89.2 |
| PointNet++ [7] | 1.48M | 90.7 |
| DGCNN [8] | 1.81M | 92.9 |
| KPConv [20] | 14.3M | 92.9 |
| RS-CNN [9] | 1.41M | 93.6 |
| DNDFN | 1.52M | 93.2 |

## 5. CONCLUSION

A deep fusion network with dual-neighborhood, *i.e.*, DNDFN, is proposed in this paper. TN-Learning and IT-Conv which can adaptively search the key neighborhood and extract rich structure information are the keys for DNDFN to achieve excellent performance. Besides, TN-Learning can also be extended to existing methods to union the points adjacent in both spactial space and feature space for each convolution region. More importantly, DNDFN not only shows strong competitiveness in idealized data but also performs well in real-world dataset, as shown in the experiments on two idealized datasets and one non-idealized datasets.